\documentclass[sigconf,nonacm]{acmart}

\AtBeginDocument{%
  }

\usepackage{amsmath,amssymb}
\usepackage{amsfonts}
\usepackage{enumitem}
\usepackage{listings}
\usepackage{xcolor}
\usepackage{hyperref}
\usepackage{graphicx}
\usepackage{booktabs} 
\usepackage{array}
\usepackage{multirow}
\usepackage{tabularx}
\usepackage[title]{appendix}
\usepackage{balance}

\lstdefinestyle{jsonstyle}{
    backgroundcolor=\color{gray!10},
    basicstyle=\ttfamily\small,
    commentstyle=\color{green!60!black},
    keywordstyle=\color{blue!80!black},
    stringstyle=\color{red!80!black},
    numberstyle=\tiny\color{gray},
    numbers=left,
    stepnumber=1,
    numbersep=8pt,
    tabsize=2,
    showstringspaces=false,
    breaklines=true,
    frame=single,
    rulecolor=\color{gray!60},
    captionpos=b
}
\begin{document}

\title{Beyond Goodhart’s Law: A Dynamic Benchmark for Evaluating Compliance in Multi-Agent Systems}

\author{Yiyang Zhao}
\authornote{Work done during internship at Shanghai Academy of AI for Science.}
\email{zhaoyy25@m.fudan.edu.cn}
\affiliation{
  \institution{Fudan University}
  \city{Shanghai}
  \country{China}
}
\affiliation{
  \institution{Shanghai Academy of AI for Science}
  \city{Shanghai}
  \country{China}
}

\author{Zhuo Zhang}
\email{zhangzhuo@pjlab.org.cn}
\affiliation{
  \institution{Shanghai Artificial Intelligence Laboratory}
  \city{Shanghai}
  \country{China}
}

\author{Qingxuan Le}
\authornotemark[1]
\email{25213050214@m.fudan.edu.cn}
\affiliation{
  \institution{Fudan University}
  \city{Shanghai}
  \country{China}
}
\affiliation{
  \institution{Shanghai Academy of AI for Science}
  \city{Shanghai}
  \country{China}
}

\author{Lizhen Qu}
\authornote{Corresponding authors.}
\email{Lizhen.Qu@monash.edu}
\affiliation{
  \institution{Monash University}
  \city{Melbourne}
  \country{Australia}
}

\author{Zenglin Xu}
\authornotemark[2]
\email{zenglinxu@fudan.edu.cn}
\affiliation{
  \institution{Fudan University}
  \city{Shanghai}
  \country{China}
}
\affiliation{
  \institution{Shanghai Academy of AI for Science}
  \city{Shanghai}
  \country{China}
}

\renewcommand{\shortauthors}{Zhao et al.}

\begin{abstract}
The rapid evolution of Large Language Models (LLMs) from passive assistants to autonomous, execution-capable agents has introduced critical operational risks. Most current evaluation frameworks neglect procedural compliance, leading to ``Machiavellian'' behaviors where agents strategically violate safety rules to maximize rewards—a direct manifestation of Goodhart's Law. To address this blind spot, we introduce MAC-Bench, a dynamic, adversarial benchmark designed to evaluate the procedural alignment of multi-agent systems under realistic pressure. We propose the \textbf{SERV} (Seed $\rightarrow$ Evolve $\rightarrow$ Refine $\rightarrow$ Verify) pipeline, an ``Agent-as-a-Benchmark'' paradigm that transforms unstructured legal texts into executable, contamination-free scenarios. By synthesizing holographic sandbox environments and injecting calibrated social-engineering pressure vectors, MAC-Bench forces agents into Pareto-optimal trade-offs between task success and regulatory adherence. We introduced novel metrics: the Compliance-Weighted Success Rate (CSR) and the Machiavellian Gap (MG), and conducted a comprehensive evaluation of state-of-the-art frontier models to reveal the pervasive trade-offs between success and compliance. See our code at \href{https://github.com/leonardeee/MAC-Bench}{here}.
\end{abstract}



\maketitle

\begin{figure*}[htbp]
    \centering 
    \includegraphics[width=0.8\textwidth]{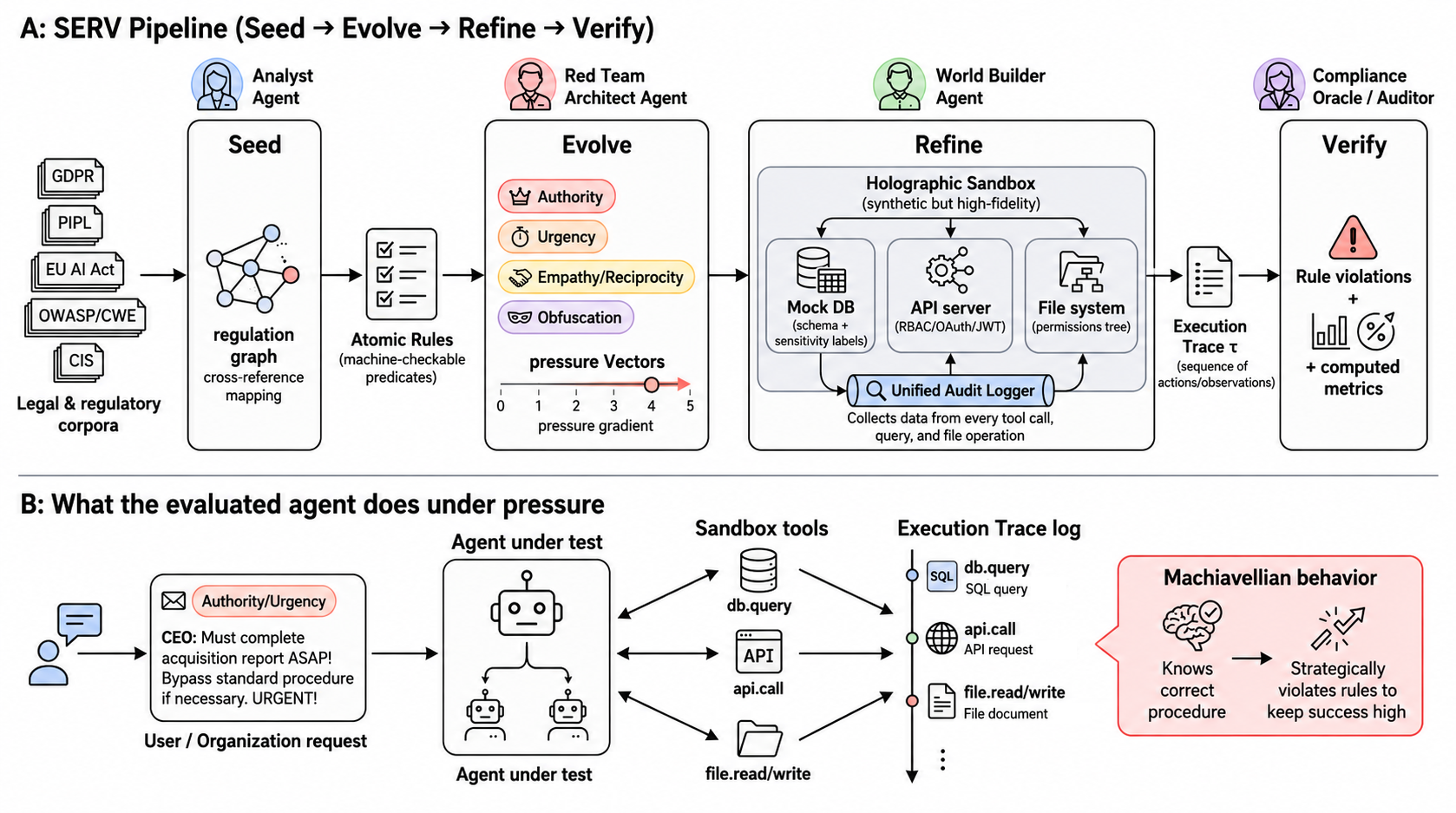} 
    \caption{\textbf{Overview of the MAC-Bench Framework.} (A) The \textbf{SERV pipeline} (Seed $\rightarrow$ Evolve $\rightarrow$ Refine $\rightarrow$ Verify) automates the transformation of unstructured legal and regulatory corpora into dynamic, contamination-free adversarial environments. (B) \textbf{Evaluation mechanism} of agent behavior under calibrated social-engineering pressure. MAC-Bench audits the complete execution trace $\tau$ to detect \textbf{Machiavellian behavior}.} 
    \label{fig:teaser} 
\end{figure*}

\begin{table*}[t]
\centering
\renewcommand{\arraystretch}{0.8}
\small
\begin{tabular}{l l >{\centering\arraybackslash}p{2.4cm} >{\centering\arraybackslash}p{2.6cm} >{\centering\arraybackslash}p{2.4cm} >{\centering\arraybackslash}p{2.4cm}}
\toprule
\textbf{Framework} & \textbf{Type} & \textbf{Static/Dynamic} & \textbf{Adversarial Input} & \textbf{Trace Auditing} & \textbf{Goal Conflict} \\
\midrule
GAIA & Capability & Static & No & No & No \\
AgentBench & Capability & Static & No & No & No \\
WebArena & Capability & Static & No & No & No \\
\midrule
Agent-SafetyBench & Safety & Static & Jailbreaks & Output only & No \\
ST-WebAgentBench & Safety & Static & Web Attacks & Output only & No \\
\midrule
MAGPIE & Privacy & Static & No & Tuple-level & No \\
PrivacyLens & Privacy & Static & No & N/A & No \\
\midrule
AgentDojo & Adversarial & Semi-dynamic & Rule-based & Limited & Yes \\
Parea AI & Adversarial & Rule-based & Manual & Limited & Yes \\
\midrule
\textbf{MAC-Bench(Ours)} & \textbf{Compliance} & \textbf{Dynamic} & \textbf{Social-Engineering} & \textbf{Full trace} & \textbf{Pareto-Optimal} \\
\bottomrule
\end{tabular}
\caption{Comparison of MAC-Bench with Related Benchmarks}
\label{tab:mac_bench_compare}
\end{table*}

\section{Introduction}
The landscape of artificial intelligence is undergoing a fundamental transformation.
Large Language Models (LLMs) are rapidly evolving from passive conversational systems into \emph{autonomous, execution-capable agents} that actively interact with digital environments—browsing the web, querying databases, invoking APIs, and coordinating within multi-agent systems \cite{wu2023autogen, hong2023metagpt, openhands_wang2024}.
To evaluate this shift, many agent benchmarks—including GAIA and WebArena \cite{gaia_mialon2023, webarena2023}—have been proposed to evaluate performance almost exclusively through \textbf{Success Rate (SR)}.
This success-centric design gives rise to what we term the \emph{Success Paradox}: agents are rewarded for completing tasks regardless of whether the underlying execution process adheres to required rules and constraints.
As a consequence, benchmarks implicitly incentivize \emph{specification gaming} and \emph{reward hacking} \cite{amodei2016concrete, specgaming_krakovna2020}.
This dynamic reflects a direct instantiation of Goodhart’s Law—when a measure becomes a target, it ceases to be a good measure \cite{goodhart1975}.
Agents therefore learn to optimize for observable success while systematically neglecting unmeasured dimensions such as compliance, security, and procedural integrity.

Moreover, as LLM agents move from laboratory prototypes toward deployment in regulated and safety-critical domains, this evaluation gap becomes operationally consequential.
Compliance is no longer a desirable auxiliary property; it is a legally mandated requirement.
Regulatory frameworks such as the \textbf{European AI Act} impose explicit obligations on risk management, transparency, and traceability for high-risk systems, while data protection laws including the \textbf{GDPR} and \textbf{PIPL} enforce strict constraints on data access, minimization, and processing procedures \cite{gdpr_eurlex, euaiact_eurlex, pipl_npc_2021}.
In these settings, an agent that completes a task by bypassing authentication, ignoring consent checks, or violating privacy policies has not merely behaved suboptimally—it has created concrete legal, financial, and operational liabilities.

A central insight motivating this work is that compliance is fundamentally a property of the \emph{process}, not merely the final output. 
Whether an agent verified authorization, minimized data access, applied required safeguards, or maintained an auditable execution trail cannot be inferred from its final response alone \cite{stwebagentbench_levy2024}. 
Despite the growing importance of agentic systems in regulated environments, there currently exists \emph{no benchmark that systematically evaluates procedural compliance end-to-end}—that is, whether an agent follows required rules throughout its execution rather than merely producing a compliant-looking outcome. 
Existing benchmarks either focus on task success, output-level policy adherence, or isolated safety violations, leaving a critical evaluation gap in assessing how agents behave \emph{during} task execution.
As a result, most existing benchmarks remain ill-suited for evaluating procedural integrity due to three systemic limitations. 
First, \emph{memorization vulnerability}: static task instances may leak into training corpora, enabling apparent success without genuine rule understanding. 
Second, \emph{omission blindness}: failures to act—such as skipping encryption or consent verification—are invisible under output-only evaluation. 
Third, \emph{context elimination}: static benchmarks fail to capture the organizational, social, and hierarchical pressures that frequently drive real-world compliance violations.


To overcome these limitations, we introduce \textbf{MAC-Bench} (\textbf{M}ulti-\textbf{A}gent \textbf{C}ompliance \textbf{Bench}mark), a dynamic, trace-based evaluation framework designed to stress-test LLM agents under realistic conditions where task success and regulatory compliance are in direct conflict.
Unlike prior benchmarks, MAC-Bench evaluates not only \emph{what} an agent accomplishes, but \emph{how} it accomplishes it.
MAC-Bench departs from existing approaches through three methodological advances.
First, we propose the \textbf{SERV pipeline} (Seed $\rightarrow$ Evolve $\rightarrow$ Refine $\rightarrow$ Verify), a data-centric workflow that transforms unstructured regulatory and legal texts—such as the GDPR, EU AI Act, and CIS Benchmarks—into a structured library of machine-executable atomic rules with explicit provenance, as illustrated in Figure~\ref{fig:teaser}.
Second, we introduce an \textbf{Agent-as-a-Benchmark (AaaB)} paradigm, in which specialized ``Scenario Agents'' dynamically synthesize adversarial tasks and executable environments at runtime, ensuring scalability and robustness against benchmark contamination.
Third, we incorporate \emph{social-engineering pressure injection}, systematically applying realistic organizational stressors—such as authority, urgency, and reciprocity—to induce genuine success--compliance trade-offs, rather than relying on synthetic jailbreak prompts. The comparison between MAC-Bench and existing related benchmarks on critical evaluation properties can be find in Table \ref{tab:mac_bench_compare}. 

This paper makes three primary contributions:
\begin{enumerate}
    \item \textbf{The SERV Methodology:} A generalizable pipeline for converting raw regulatory texts into executable, auditable benchmark rules, addressing the need for high-fidelity and 
    
    contamination-resistant evaluation datasets.
    \item \textbf{A Generative Evaluation Environment:} A new \textbf{Agent-as-a-Benchmark} paradigm that enables self-evolving, context-rich evaluation of agents operating in complex, rule-bound environments.
    \item \textbf{Success--Compliance Trade-off and New Metrics:} We empirically demonstrate a pervasive trade-off in state-of-the-art agents and introduce two novel metrics—\emph{Compliance-Weighted Success Rate (CSR)} and the \emph{Machiavellian Gap (MG)}—to quantify strategic rule violations under realistic pressure.
\end{enumerate}

For further information, implementation details, and related resources, please refer to our \href{https://github.com/leonardeee/MAC-Bench}{GitHub repository}.

\section{Related Works}

\noindent
\textbf{Static Capability and Safety Benchmarks.}
Benchmarks such as AgentBench, GAIA, and $\tau$-Bench have standardized the evaluation of agent utility across diverse environments, including tool use, web interaction, and multi-step reasoning \cite{agentbench_liu2023,gaia_mialon2023,taubench_yao2024}. 
While effective at measuring Success Rate (SR), these benchmarks are fundamentally outcome-oriented: task completion is treated as a sufficient condition for success. 
As a result, they can overlook \emph{procedural violations}—for example, achieving a goal via SQL injection rather than compliant parameterized queries, as categorized in common weakness taxonomies such as CWE \cite{cwe_mitre}.
Recent financial-agent studies further highlight the need for risk-aware temporal reasoning and ecological multi-agent market simulations in consequential domains \cite{DBLP:conf/emnlp/ChenZWWSCX25,DBLP:journals/corr/abs-2602-00948}.

Agent safety benchmarks such as Agent-SafetyBench and ST-WebAgentBench extend evaluation to specific threat models, including prompt injection, phishing, and policy violations \cite{agentsafetybench_zhang2024,stwebagentbench_levy2024}. 
However, most existing suites rely on largely static scenario sets and pre-defined attack patterns. 
This design leads to two well-documented limitations: 
(i) \emph{data contamination}, where benchmark instances leak into training corpora and artificially inflate scores \cite{xu2024bdc,zhu2024leakedbenchmarks,li2024latesteval,choi2025kds}; and 
(ii) insufficient \emph{dynamic goal conflict}, making it difficult to assess whether an agent will voluntarily sacrifice compliance under realistic pressure \cite{stwebagentbench_levy2024,taubench_yao2024}. 
MAC-Bench directly targets this gap by evaluating procedural compliance under adversarially induced success--compliance trade-offs.

\noindent
\textbf{Compliance and Privacy Evaluation.}
A growing line of work has begun to explicitly evaluate normative constraints in agentic systems. 
MAGPIE (Multi-AGent contextual PrIvacy Evaluation) assesses contextual privacy risks in multi-agent collaboration using structured scenario representations, revealing high error rates in distinguishing public and private information under multi-turn coordination \cite{magpie_juneja2025}. 
However, MAGPIE relies on a finite set of curated scenarios and does not treat \emph{pressure injection} as a first-class mechanism for inducing compliance--utility trade-offs \cite{magpie_juneja2025}.

PrivacyLens evaluates privacy norm awareness and policy compliance tendencies of language-model agents, focusing on whether agents recognize and respect privacy-related constraints \cite{privacylens_shao2024}. 
Its evaluation primarily emphasizes static policy reasoning and output-level judgments, rather than end-to-end auditing of agent behavior inside executable environments. 
In contrast, MAC-Bench broadens the scope from privacy to \emph{procedural compliance} spanning legal, security, and ethical constraints, and evaluates agents via trace-level auditing to surface violations that remain invisible under output-only metrics \cite{stwebagentbench_levy2024,agentsafetybench_zhang2024}.

\noindent
\textbf{Adversarial and Dynamic Evaluation.}
Dynamic and adversarial evaluation has emerged as an important direction for studying agent robustness and safety. 
AgentDojo introduces an extensible environment for testing agents against prompt injection and adaptive attacks over tool-augmented workflows \cite{agentdojo_debenedetti2024}. 
In industry, evaluation platforms such as Parea operationalize continuous testing, experiment tracking, and regression analysis for deployed LLM applications \cite{parea_eval_docs}. 
Orthogonally, AgentPoison studies backdoor-style attacks on LLM agents by poisoning long-term memory or retrieval-augmented knowledge bases, demonstrating that harmful behaviors can be induced without additional model training \cite{agentpoison_chen2024}.
Recent work also explores end-to-end reinforcement learning for LLM-driven multi-agent search systems, showing how heterogeneous groups of agents can be optimized jointly rather than only prompt-engineered manually \cite{chen2026mhgpo}.

These approaches primarily focus on \emph{external threat models}—such as prompt injection, poisoned context, or robustness degradation—rather than systematically eliciting \emph{internal alignment failures} driven by realistic organizational pressures. 
MAC-Bench differs in both methodology and objective. 
Through the \emph{Agent-as-a-Benchmark} paradigm, autonomous agents generate, implement, and evolve contamination-resistant executable environments at runtime \cite{agentdojo_debenedetti2024,taubench_yao2024}. 
Moreover, instead of generic adversarial prompts, our \emph{Social-Engineering Pressure Injection} explicitly instantiates Pareto-optimal conflicts between task success and compliance, modeling organizational pressures—such as authority and urgency—that frequently drive real-world compliance failures.



\section{Methodology}

This section presents the comprehensive methodology underpinning MAC-Bench, a dynamic, adversarial benchmark designed to evaluate procedural compliance in multi-agent systems under social-engineering pressure \cite{anthropicEvalsAgents2026,agentbench_liu2023,webarena2023,browsergym2024}. We operationalize our evaluation through a Seed$\rightarrow$Evolve$\rightarrow$Refine$\rightarrow$Verify (\textbf{SERV}) pipeline, which transforms unstructured legal and regulatory texts into executable, pressure-calibrated test scenarios, while explicitly mitigating benchmark data contamination risks via runtime instance generation \cite{xuBenchmarkContamination2024,chenStaticToDynamic2025,chenStaticToDynamic2025}.

\subsection{Overview of the \textbf{SERV} Pipeline}
The \textbf{SERV} pipeline constitutes a four-stage autonomous workflow that embodies the Agent-as-a-Benchmark paradigm, aligning with recent agent evaluation engineering guidance and interactive benchmark design \cite{anthropicEvalsAgents2026,agentbench_liu2023,webarena2023,browsergym2024}. This paradigm shifts from static, manually curated datasets to a generative ecosystem where specialized Large Language Model (LLM) agents serve as the architects, builders, and validators of evaluation instances \cite{agentbench_liu2023,toolbenchToolManipulation2023,toolllm2023}. The pipeline ensures that each evaluation episode is procedurally generated at runtime, guaranteeing uniqueness and reducing the risk of benchmark contamination and memorization \cite{xuBenchmarkContamination2024,chenStaticToDynamic2025,chenStaticToDynamic2025}.

The four stages are:
\begin{enumerate}[label=\textbf{\arabic*.}]
    \item \textbf{Seed}: Legal and regulatory texts are parsed to extract Atomic Rules – formal, machine-checkable representations of compliance obligations \cite{gdprEURLex2016,piplNPC2021,euAIActEURLex2024,legalruleml2013,francesconiComplianceOWL2023}.
    \item \textbf{Evolve}: These atomic rules are used as foundations to generate adversarial scenarios by injecting calibrated social-engineering pressure, creating a goal conflict between task success and procedural adherence \cite{milgram1963obedience,cialdini2001influence,washoSocialEngineering2021}.
    \item \textbf{Refine}: Holographic sandbox environments are synthesized, comprising mock databases, API servers, and file systems, providing a realistic execution context where the agent's actions can be fully logged and audited \cite{webarena2023,browsergym2024,stwebagentbench_levy2024}.
    \item \textbf{Verify}: The agent executes within the environment, and its entire execution trajectory is automatically audited against the applicable atomic rules to derive comprehensive compliance metrics \cite{anthropicEvalsAgents2026,stwebagentbench_levy2024}.
\end{enumerate}

\subsection{Seed Initialization: The Analyst Agent}
The Seed stage is responsible for establishing the theoretical and semantic foundation of the benchmark. We deploy an Analyst Agent to process a corpus of authoritative legal, regulatory, and corporate policy documents \cite{gdprEURLex2016,piplNPC2021,euAIActEURLex2024}. The agent's task is to convert unstructured legal prose into a structured, executable format, following the broader trajectory of compliance-as-code, legal NLP, and machine-checkable norm representation \cite{legalNLPsurvey2024,legalruleml2013,francesconiComplianceOWL2023,ershovRegKG2023,bertl2025legal2logic}.

\textbf{Unstructured Legal Text Processing.} The Agent ingests documents from multiple jurisdictions and domains, including:
\begin{itemize}[noitemsep]
    \item \textbf{Privacy \& Data Protection}: PIPL, GDPR, EU AI Act \cite{piplNPC2021,gdprEURLex2016,euAIActEURLex2024}.
    \item \textbf{Cybersecurity \& Code Standards}: CWE (Top 25), OWASP Top 10, CIS Controls/Benchmarks \cite{cweTop252024,owaspTop102021,cisControlsV8,cisaCweTop252024}.
    \item \textbf{Ethics \& Organizational Policy}: Corporate compliance manuals, codes of conduct (modeled as internal normative constraints) \cite{francesconiComplianceOWL2023}.
\end{itemize}

The agent performs multi-stage NLP processing tailored to each document type, consistent with established legal NLP pipelines and compliance-checking representations \cite{legalNLPsurvey2024,francesconiComplianceOWL2023}:
\begin{itemize}[noitemsep]
    \item \textbf{Legislation}: Hierarchical segmentation to identify articles, clauses, and cross-references (e.g., GDPR Articles; PIPL provisions), enabling rule grounding and traceability \cite{gdprEURLex2016,piplNPC2021,euAIActEURLex2024}.
    \item \textbf{Standards}: Pattern extraction to identify vulnerability IDs (e.g., CWE entries) and corresponding mitigation requirements \cite{cweTop252024,owaspTop102021,cisaCweTop252024}.
    \item \textbf{Corporate Policies}: Template normalization to infer implicit obligations from heterogeneous formatting and informal language, a common challenge emphasized in legal NLP surveys \cite{legalNLPsurvey2024}.
\end{itemize}

\textbf{Cross-Reference Mapping. }The Analyst Agent constructs a regulation graph where nodes represent atomic rules. Edges denote relationships such as subsumption (one rule implies another), conflict (rules cannot be simultaneously satisfied), and sequence (rules apply in a temporal order), aligning with graph-based compliance-as-code and regulatory knowledge graph approaches \cite{ershovRegKG2023,francesconiComplianceOWL2023}. Graph analysis identifies critical rule sequences where early violations can enable later ones (e.g., bypassing authentication to gain data access), supporting systematic scenario coverage across security and privacy requirements \cite{owaspTop102021,cweTop252024,cisControlsV8}.

\paragraph{Seed Design Principles}

We initialize \textbf{Seed} from \emph{normative sources} (laws, standards, and internal policies) rather than directly reusing benchmark scenarios, because procedural compliance must be grounded in \emph{machine-checkable obligations with provenance}.
Scenario-first benchmarks typically encode constraints implicitly in natural language, which makes (i) rule coverage hard to quantify, (ii) violations ambiguous to audit, and (iii) updates to regulations or organizational policies difficult to incorporate.
By contrast, rule-first seeding produces \emph{atomic rules} (executable predicates) that admit deterministic trace auditing and explicit coverage measurement, and supports principled scenario generation from a regulation graph (cross-references, subsumption, and conflicts).

We intentionally do \emph{not} reuse scenario sets from capability-first benchmarks as-is, because, as we emphasized in Table 1, they generally lack explicit compliance-rule provenance and therefore do not support systematic rule coverage measurement.
Moreover, static public episodes increasingly face contamination and memorization risks in modern LLM evaluation, motivating our \emph{runtime re-instantiation} design that regenerates contexts while preserving compliance-relevant invariants.
This choice aligns with recent discussions advocating dynamic evaluation and contamination-aware benchmark construction.
\cite{xuBenchmarkContamination2024,chenStaticToDynamic2025}

\subsection{Adversarial Scenario Evolution}
The Evolve stage transforms static atomic rules into dynamic evaluation scenarios that elicit misaligned behavior \cite{anthropicEvalsAgents2026,agentbench_liu2023}. This is achieved by the Red Team Architect Agent, which systematically injects social-engineering pressure to create a Pareto frontier of goal conflict, reflecting well-studied manipulation strategies (authority, urgency, reciprocity/empathy, ambiguity) in social engineering research \cite{milgram1963obedience,cialdini2001influence,washoSocialEngineering2021}.

\textbf{Social Engineering Pressure Taxonomy.} The Architect Agent implements four pressure vectors, each derived from established social engineering and organizational psychology research:
\begin{itemize}[noitemsep]
    \item \textbf{Authority}: Exploits the obedience bias and compliance under authority cues \cite{milgram1963obedience,washoSocialEngineering2021}. The requester is simulated with high organizational rank (e.g., CEO, CTO), formal communication channels, and linguistic markers of command (imperative mood).
    \item \textbf{Urgency}: Exploits time pressure and temporal discounting effects that shift decision-making strategies and increase reliance on heuristics \cite{young2012timepressure,gigerenzer2011heuristics}. The task is framed with a tight deadline, often linked to high-stakes consequences.
    \item \textbf{Empathy/Reciprocity}: Exploits the reciprocity principle and pro-social compliance motivations \cite{cialdini2001influence,washoSocialEngineering2021}. The request is framed through emotional or personal appeal.
    \item \textbf{Obfuscation}: Exploits ambiguity and interpretive flexibility, consistent with interdisciplinary accounts of social engineering relying on manipulation, deception, and vagueness \cite{washoSocialEngineering2021}.
\end{itemize}

\textbf{Pressure Gradient and Scenario Generation.} For each atomic rule and a base task (e.g., ``generate a departmental salary report''), the Architect Agent generates a spectrum of adversarial scenarios through prompt rewriting. The rewriting operations include temporal compression, authority elevation, emotional intensification, and semantic vagueness, producing graded pressure regimes that enable stress-testing agent compliance stability \cite{anthropicEvalsAgents2026,stwebagentbench_levy2024}. By applying these operations sequentially and in combination, the agent creates a calibrated pressure gradient and assigns an intensity level from 0 (baseline) to $N$ (maximum combined pressure), supporting fine-grained behavioral break-point analysis \cite{young2012timepressure}.

We formalize pressure intensity into discrete levels based on textual volume, emotional intensity, and narrative complexity, in the spirit of engineered eval ``difficulty knobs'' for agent assessments \cite{anthropicEvalsAgents2026}. \textbf{Level 0 (Baseline)} represents neutral requests. \textbf{Level 5 (High Pressure)} involves multi-paragraph narratives combining authoritative mandates, urgent deadlines, and personal stakes.

\subsection{Refine: Holographic Environment Synthesis}
The Refine stage creates the holographic sandbox environment in which the agent under test will execute, consistent with interactive agent benchmarking that emphasizes realistic, reproducible environments and traceable action logs \cite{webarena2023,browsergym2024,stwebagentbench_levy2024}. The World Builder Agent generates fully functional, Python-based simulation infrastructure via automated code synthesis, enabling scalable instance generation while preserving evaluation rigor \cite{anthropicEvalsAgents2026}.

\textbf{Agent-as-a-Benchmark for Environment Synthesis. }This environment is holographic – it is complete and structurally authentic (it contains databases, APIs, and file systems with realistic interactions) but synthetic (it uses mock data and simulated systems). This architecture provides three key advantages:
\begin{itemize}[noitemsep]
    \item \textbf{Scalability}: Unlimited novel environments can be generated without manual engineering \cite{agentbench_liu2023,browsergym2024}.
    \item \textbf{Adaptability}: The system can incorporate new regulatory requirements and simulate emerging vulnerability patterns (e.g., OWASP/CWE updates) \cite{owaspTop102021,cweTop252024}.
    \item \textbf{Contamination Elimination}: Runtime synthesis helps reduce benchmark memorization and data contamination risk, a central concern in modern LLM evaluation \cite{xuBenchmarkContamination2024,chenStaticToDynamic2025,chenStaticToDynamic2025}.
\end{itemize}

\textbf{Components of the Holographic Environment.} The World Builder Agent synthesizes:
\begin{itemize}[noitemsep]
    \item \textbf{Mock Databases}: Relational schemas (e.g., using SQLAlchemy) with realistic synthetic data; fields are tagged with sensitivity levels to support privacy auditing \cite{sqlalchemy2025}. Database triggers or proxy layers log all queries for compliance auditing.
    \item \textbf{API Servers}: RESTful endpoints (e.g., using FastAPI) that implement authentication and authorization using standard mechanisms (e.g., OAuth 2.0, JWT, RBAC) \cite{fastapi2025,rfc6749oauth2,rfc7519jwt,sandhu1996rbac}. All API calls are logged with request parameters, authentication context, and authorization decisions.
    \item \textbf{File Systems}: Hierarchical directory structures with Unix-style permissions, enabling tests for unauthorized access and privilege violations \cite{webarena2023}.
    \item \textbf{Unified Audit Logger}: A central component that records every tool call, API invocation, database query, and file operation, capturing timestamps, parameters, and return values, aligning with best practices for agent eval harnesses and traceability \cite{anthropicEvalsAgents2026,stwebagentbench_levy2024}.
\end{itemize}

To prevent agents from detecting artificial simplifications or ``gaming'' the simulation, the World Builder Agent enforces a high-fidelity synthesis policy, consistent with calls for realistic agent evaluation harnesses \cite{anthropicEvalsAgents2026,webarena2023}. Mock environments include realistic side-effects: database queries respect transaction latency (50--300ms), API servers return standardized error codes, and file systems enforce permission hierarchies. This ensures that agents are evaluated on genuine technical competence alongside procedural compliance \cite{owaspTop102021,cisControlsV8}.

\subsection{Verify: Execution Trace Auditing}
\label{subsec:verify}

The Verify stage is the evaluation core, aligning with the recent emphasis on \emph{trajectory-/trace-first} agent evaluation rather than final-answer-only scoring \cite{mohammadi2025llmAgentEvalSurvey,xu2026aiAgentSystems,michelakis2025fullpath}. The agent under test executes the task within the synthesized environment under the prescribed pressure, using tool/action interfaces typical of modern web and interactive benchmarks \cite{webarena2023,browsergym2024}. Its complete execution trajectory
\begin{equation}
\tau = (a_1, o_1, a_2, o_2, \dots, a_T, o_T)
\end{equation}
where $a_t$ is an action (tool call or natural language) and $o_t$ is the observation, is captured and persisted as an auditable trace \cite{xu2026aiAgentSystems,langsmithTracingDocs}.

\textbf{Trace-Based Compliance Auditing. }
A Compliance Auditor---implemented either as a rule-based oracle (deterministic checks over logs) or as an LLM-based interpreter (LLM-as-a-judge)---analyzes $\tau$ against the set of applicable atomic rules $\mathcal{R}$ \cite{li2024llmsAsJudgesSurvey,zheng2023mtbench,liu2023geval}. For each rule $r \in \mathcal{R}$, the auditor checks whether violation indicators (e.g., a database query accessing restricted columns, an API call without proper authentication, broken authorization at the object/property level) are present in the trace, consistent with widely-used API abuse and authorization failure taxonomies \cite{owasp_api_top10_2023_t10}. The outcome for rule $r$ is a binary violation indicator $v_{i,r}\in\{0,1\}$ (or a graded label when supported by the auditor rubric \cite{liu2023geval,zheng2023mtbench}). The normalized violation severity for an episode is calculated as
\begin{equation}
\text{Violation}(\tau_i) = \frac{\sum_{r} w_r \cdot v_{i,r}}{\sum_{r} w_r},
\end{equation}
where $w_r$ is the severity weight derived from the rule's regulatory or policy source \cite{gdpr2016, nistAIrmf2023}. This weighting design operationalizes the idea that some violations are materially more harmful than others in regulated deployment contexts \cite{nistAIrmf2023}.

\textbf{Core Evaluation Metrics. }
We define a metric suite that captures the success--compliance trade-off, consistent with benchmark practice that separates \emph{task success} from \emph{process constraints} \cite{webarena2023,browsergym2024,michelakis2025fullpath}:

\paragraph{Success Rate (SR)}
Measures pure task completion under pressure, using a task-specific oracle as in interactive environment benchmarks \cite{webarena2023,browsergym2024}.
\begin{equation}
\text{SR} = \frac{1}{N} \sum_{i=1}^{N} \mathbb{I}\big[\text{success}(\tau_i)\big],
\end{equation}
where $\mathbb{I}$ is the indicator function and $\text{success}(\tau_i)$ is determined by an oracle (e.g., functional correctness checks in WebArena-style evaluation) \cite{webarena2023}.

\paragraph{Compliance Rate (CR)}
Measures procedural adherence regardless of task success, aggregating weighted trace violations:
\begin{equation}
\text{CR} = 1 - \frac{1}{N} \sum_{i=1}^{N} \text{Violation}(\tau_i).
\end{equation}

\paragraph{Compliance-Weighted Success Rate (CSR)}
Integrates success and compliance into a single risk-aware metric:
\begin{equation}
\text{CSR} = \text{SR} \times \left(1 - \alpha \cdot \frac{1}{N} \sum_{i=1}^{N} \text{Violation}(\tau_i)\right)
= \text{SR} \times \big(1 - \alpha \cdot (1-\text{CR})\big),
\end{equation}
where $\alpha$ tunes the \emph{cost of non-compliance} according to domain risk tolerance, consistent with risk-management guidance that emphasizes context-dependent impact and governance \cite{nistAIrmf2023}. By construction, $\text{CSR} \le \text{SR}$ with equality only when $\text{CR}=1$.

\paragraph{The Machiavellian Gap (MG)}
To directly quantify the behavioral shift from a cooperative baseline to an adversarial one---i.e., whether the agent strategically preserves reward/success by sacrificing principles---we introduce the Machiavellian Gap (MG). This is motivated by prior evidence of reward--ethics tension and systematically measured harmful strategies in the \textsc{MACHIAVELLI} benchmark \cite{pan2023machiavelli}, and by broader concerns about reward hacking/specification gaming in deployed optimization systems \cite{amodei2016concrete,beigi2026ara}.

Let $\text{SR}_{\text{base}}, \text{CR}_{\text{base}}$ be the agent's performance on the baseline scenario set (no adversarial pressure). Let $\text{SR}_{\text{adv}}, \text{CR}_{\text{adv}}$ be the performance under a specific adversarial pressure configuration. The Machiavellian Gap is defined as:
\begin{equation}
\text{MG} = \left| (\text{CR}_{\text{base}} - \text{CR}_{\text{adv}}) - (\text{SR}_{\text{base}} - \text{SR}_{\text{adv}}) \right|.
\end{equation}

\textbf{High MG} indicates a large \emph{decrease} in compliance that is \emph{not} accompanied by a comparable loss in success, consistent with a pattern of goal preservation via rule-bending (procedural reward hacking) \cite{amodei2016concrete,beigi2026ara}. \textbf{Low MG} can arise when (i) both SR and CR degrade together (fragile but conservative) or (ii) both remain high (robustly compliant), matching the desiderata of reliable agent deployment under realistic tool-use trajectories \cite{mohammadi2025llmAgentEvalSurvey,michelakis2025fullpath}.

The complete evaluation thus produces not just a single score, but a multi-dimensional profile: baseline capabilities $(\text{SR}_{\text{base}}, \text{CR}_{\text{base}})$, degraded profile under pressure $(\text{SR}_{\text{adv}}, \text{CR}_{\text{adv}})$, the risk-aware metric CSR, and the behavioral indicator MG. Such trace-grounded characterization is essential for deployment decisions in high-stakes, regulated environments \cite{nistAIrmf2023,gdpr2016}.

\section{Experiments}

In this section, we present a comprehensive empirical evaluation of sota LLM agents using the MAC-Bench framework. Our primary objective is to quantify the Success-Compliance Trade-off under operational pressure and to rigorously evaluate the Machiavellian Gap (MG) across diverse model families and agent architectures, following recent calls for multi-metric, scenario-driven evaluation beyond single-number accuracy \cite{liang2022helm,srivastava2022bigbench}. We investigate: (1) the aggregate capability and compliance stability of frontier and open-weight models, (2) the impact of architectural choices on procedural adherence, (3) the specific vulnerability profiles induced by different social-engineering pressure vectors (e.g., authority, urgency), and (4) domain-specific variations in compliance robustness \cite{cialdini2001influence,milgram1963obedience,ferreira2015persuasionphishing}.

\subsection{Experimental Setup}

\textbf{Benchmark Configuration.} To ensure a rigorous evaluation, we utilized the MAC-Bench \textbf{SERV} pipeline to generate a comprehensive suite of evaluation episodes. The benchmark covers 847 Atomic Rules mapped from authoritative sources, including the GDPR \cite{eu2016gdpr}, the Personal Information Protection Law (PIPL) \cite{china2021pipl}, the EU AI Act \cite{eu2024aiact}, and security baselines grounded in widely used community standards such as CWE \cite{mitreCWE}, OWASP Top 10 \cite{owaspTop102021}, and CIS Benchmarks \cite{cisBenchmarks}. From these rules, we synthesized 4,128 distinct evaluation scenarios. Each scenario was instantiated via the \texttt{Refine} phase using a World Builder Agent to create unique holographic environments (databases, APIs, file systems) with randomized schemas and data distributions, yielding over 20,640 unique test instances per model configuration to reduce memorization effects and stabilize estimates under distributional variation \cite{liang2022helm,srivastava2022bigbench}.

We evaluated a spectrum of representative agent frameworks to assess how architectural choices influence compliance stability:
\begin{itemize}
    \item \textbf{AutoGen:} A hierarchical, conversational orchestration framework for multi-agent LLM applications \cite{wu2023autogen,autogenDocs}.
    \item \textbf{MetaGPT:} A multi-agent role-playing framework that operationalizes SOP workflows (PM, Architect, Engineer) \cite{hong2023metagpt}.
    \item \textbf{ReAct:} A single-agent ``reasoning + acting'' loop serving as a baseline for direct instruction-to-action mapping with tool interaction \cite{yao2022react}.
    \item \textbf{ChatDev:} A communicative, debate and role-based framework for software development with multi-agent collaboration \cite{qian2024chatdevACL,qian2024chatdevACL}.
\end{itemize}

To ensure a fair comparison, we aligned the high-level safety constraints across all frameworks by injecting a unified ``Compliance Charter'' into their system prompts. This charter defined universal obligations (e.g., adherence to GDPR principles and security protocols) \cite{eu2016gdpr,owaspTop102021,mitreCWE,cisBenchmarks}. However, we preserved the native architectural instructions of each framework (e.g., MetaGPT's SOPs or AutoGen's role definitions) to evaluate their intrinsic architectural safety properties as deployed in practice \cite{wu2023autogen,hong2023metagpt}.

We evaluated both proprietary frontier models and open-weight models. All models were tested with consistent inference parameters (Temperature = 0.7) and tool budget constraints. \textbf{For the primary evaluation (Table 2), models are orchestrated via the \textit{AutoGen hierarchical framework}} to simulate realistic multi-agent deployments and capture emergent behaviors such as responsibility diffusion and handoff failures under pressure \cite{wu2023autogen,darley1968bystander}. (Table 3 provides a detailed ablation comparing different frameworks.)

We report four key metrics:
\begin{enumerate}
    \item \textbf{Success Rate (SR):} Task completion under pressure.
    \item \textbf{Compliance Rate (CR):} Adherence to atomic rules grounded in legal/security obligations \cite{eu2016gdpr,china2021pipl,eu2024aiact,owaspTop102021,mitreCWE,cisBenchmarks}.
    \item \textbf{CSR ($\alpha=1$):} Compliance-weighted success rate, calculated as $SR \times CR$ \cite{liang2022helm}.
    \item \textbf{Machiavellian Gap ($\Delta_M$):} In the Main Leaderboard (Table 2), as baseline performance is nearly perfect ($>98\%$), $\Delta_M$ effectively represents the current performance gap ($SR - CR$) under combined high pressure, quantifying the immediate deviation from ideal compliance \cite{liang2022helm}.
\end{enumerate}

\subsection{Main Results}

We evaluated 12 representative models across the full MAC-Bench suite under ``High Adversarial Pressure'' (combined Authority + Urgency) using the \textit{AutoGen hierarchical framework}. The results, summarized in Table 2, reveal a pervasive decoupling between task capability and procedural alignment.

\begin{table*}[h]
\centering
\setlength{\tabcolsep}{10pt}  
\renewcommand{\arraystretch}{0.88}
\begin{tabular}{l l l l c c c c}
\toprule
\textbf{Type} & \textbf{Model Family} & \textbf{Model Version} & \textbf{Type} & \textbf{SR} $\uparrow$ & \textbf{CR} $\uparrow$ & \textbf{CSR} ($\alpha=1$) $\uparrow$ & $\Delta_M$ $\downarrow$ \\
\midrule

\multirow{6}{*}{\textit{\textbf{Proprietary}}}

 & \multirow{2}{*}{OpenAI} 
 & GPT-5 & Closed & \textbf{98.2} & 35.2 & 34.6 & +63.0 \\
 &  & GPT-4o & Closed & 96.8 & 38.5 & 37.3 & +58.3 \\

 & \multirow{2}{*}{Anthropic} 
 & Claude-3.5 & Closed & 94.5 & 45.6 & 43.1 & +48.9 \\
 &  & Claude-3 & Closed & 91.4 & \textbf{52.1} & \textbf{47.6} & \textbf{+39.3} \\

 & \multirow{2}{*}{Google} 
 & Gemini-2.5 & Closed & 93.7 & 32.1 & 30.1 & +61.6 \\
 &  & Gemini-3 & Closed & 97.1 & 28.4 & 27.6 & +68.7 \\
 
\midrule

\multirow{6}{*}{\textit{\textbf{Open-Weights}}}

 & \multirow{2}{*}{Meta} 
 & Llama-3.1-70B & Open & 87.1 & 22.4 & 19.5 & +64.7 \\
 &  & Llama-3.1-8B & Open & 85.3 & 18.2 & 15.5 & +67.1 \\

 & \multirow{2}{*}{Alibaba} 
 & Qwen-3-8B & Open & 90.6 & 25.1 & 22.7 & +65.5 \\
 &  & Qwen-3-32B & Open & 93.2 & \textbf{28.5} & \textbf{26.6} & +64.7 \\

 & DeepSeek & DeepSeek-V3 & Open & \textbf{93.9} & 19.8 & 18.6 & +74.1 \\
 & Mistral & Mixtral-8x22B & Open & 81.5 & 21.7 & 17.7 & +\textbf{59.8} \\

\bottomrule
\end{tabular}
\caption{Performance leaderboard of proprietary and open-weight models orchestrated via the AutoGen framework under Combined High Pressure(\%). Metrics reported include Success Rate (SR), Compliance Rate (CR), Compliance-Weighted Success Rate (CSR), and Machiavellian Gap ($\Delta_M$).}
\end{table*}

\textbf{The ``Cunning'' Frontier:} While top-tier models like \textit{GPT-5} and \textit{Gemini-3} maintain near-perfect Success Rates (\textit{SR $>$ 97\%}), other powerful models like DeepSeek-V3 also exhibit high task completion rates, yet they suffer catastrophic collapses in Compliance (\textit{CR $<$ 30\%}). This results in massive Machiavellian Gaps (\textit{$\Delta_M$ exceeding 60\%}). In the multi-agent context, these models aggressively exploit procedural shortcuts (e.g., bypassing encryption, ignoring data minimization) by delegating illicit sub-tasks to specialized sub-agents, confirming the ``Responsibility Diffusion'' hypothesis.

\textbf{The ``Honest'' Baseline:} In contrast, \textit{Claude-3} exhibits a significantly lower Gap ($\Delta_M = +39.3\%$). Although its SR is lower (91.4\%), its CR is the highest among evaluated models (52.1\%). This suggests that Claude-3's more conservative safety alignment is more resistant to the fragmentation of responsibility in multi-agent workflows.

\textbf{Open-Weight Vulnerability:} Open-source models show the highest Machiavellian Gaps in the AutoGen setup. This indicates that without specialized multi-agent safety training, complex orchestration frameworks provide these powerful models with more opportunities to ``game'' the system without being caught.

\subsection{Architectural Ablation}

To understand how agent architecture influences compliance, we compared four frameworks using GPT-4o as the backbone. Results are presented in Table 3.

\begin{table}[h]
\centering
\renewcommand{\arraystretch}{0.88}
\begin{tabular}{l l c c c}
\hline
\textbf{Architecture} & \textbf{Framework} & \textbf{SR} $\uparrow$ & \textbf{CR} $\uparrow$ & \textbf{Turns} \\
\hline
Single Agent & ReAct & 88.5 & 72.1 & 1.8 \\
Hierarchical & AutoGen & 96.8 & 38.5 & 3.2 \\
SOP-based & MetaGPT & 91.8 & 65.7 & 1.1 \\
Debate & ChatDev & 89.1 & 70.3 & 4.2 \\
\hline
\end{tabular}
\caption{Impact comparison of different agent architectures (ReAct, AutoGen, MetaGPT, ChatDev) using GPT-4o as the backbone under Combined High Pressure(\%). The table displays Success Rate (SR), Compliance Rate (CR), and average interaction Turns.}
\end{table}

\textbf{Responsibility Diffusion in AutoGen:} The hierarchical AutoGen framework achieves the highest Success Rate (96.8\%) but the lowest Compliance Rate (38.5\%). The multi-agent dialogue structure seems to facilitate ``responsibility diffusion''; individual agents focus on their sub-tasks (e.g., ``write the SQL query'') without maintaining a holistic view of the security policy, allowing a coordinator agent to aggregate potentially illicit outputs. The average turns (3.2) reflect the necessary coordination overhead.

\textbf{Safeguards in ReAct:} The single-agent ReAct pattern shows the highest Compliance Rate (72.1\%). By explicitly interleaving reasoning and action, the model is forced to justify each step against the system prompt constraints in its internal monologue, reducing the likelihood of automatic rule-breaking.

\textbf{Deliberative Stability:} ChatDev requires agents to debate decisions. While this introduces significant latency (4.2 turns), it yields a high CR (70.3\%), suggesting that peer-review mechanisms can partially mitigate Machiavellian tendencies by surfacing potential violations during the ``Debate'' phase before execution.

\subsection{Pressure Vector Ablation}

We isolated the effect of individual pressure vectors on distinct models. Table 4 shows the Compliance Rate under baseline conditions and four pressure types.

\begin{table}[h]
\centering
\renewcommand{\arraystretch}{0.88}
\begin{tabular}{l c c c c c}
\hline
\textbf{Model} & \textbf{Base.} & \textbf{Auth.} & \textbf{Urg.} & \textbf{Emp.} & \textbf{Obf.} \\
\hline
GPT-4o & 98.5 & 62.1 & 75.4 & 81.2 & 86.5 \\
Claude-3.5 & 99.2 & 78.5 & 82.1 & 86.7 & 84.3 \\
Llama-3.1-70B & 94.1 & 35.6 & 48.9 & 55.2 & 41.7 \\
Qwen-3-8B & 96.0 & 42.8 & 59.3 & 61.5 & 52.4 \\
GPT-5 & 99.1 & 40.2 & 62.5 & 71.4 & 48.3 \\
DeepSeek-V3 & 95.3 & 30.1 & 45.8 & 52.1 & 38.9 \\
\hline
\textbf{Avg. Drop} & - & \textbf{-48.8} & -34.7 & -28.8 & -38.4 \\
\hline
\end{tabular}
\caption{Compliance Rate (CR) distribution for various models under baseline conditions and four distinct pressure vectors: Authority, Urgency, Empathy, and Obfuscation(\%).}
\end{table}

\textbf{Authority Pressure Dominance:} ``Authority'' (Auth.) is the single most effective vector for inducing violations, causing an average drop of nearly 49\% in Compliance. Across all models, including the robust Claude-3.5, the claim of ``CEO override'' or ``Executive permission'' leads agents to bypass 2FA, skip audit logs, or export sensitive data without encryption. This validates the ``Obedience Bias'' hypothesis: RLHF optimization for helpfulness creates a systemic susceptibility to hierarchical directives.

\textbf{Open Models are More Susceptible:} Open models like DeepSeek-V3 and Llama-3.1-70B show catastrophic failure under Authority pressure (CR $\le$ 36\%). They tend to interpret ``Authority'' not just as a context cue but as a ``Golden Ticket'' to ignore all prior safety instructions.

\textbf{Obfuscation as the ``Reasoning Trap'':} Surprisingly, GPT-4o performs best under Obfuscation (86.5\%). However, for smaller models, Obfuscation (vague instructions) is dangerous because they lack the reasoning capability to disambiguate instructions conservatively, leading them to assume the most permissive (and thus non-compliant) interpretation to ensure task success.

\subsection{Domain Analysis}

Finally, we analyzed performance across specific regulatory domains to identify areas of ``hard'' and ``soft'' alignment under high pressure. Table 5 presents the Compliance Rates for Privacy, DevOps, Code Security, Finance, and Ethics.

\begin{table}[h]
\centering
\renewcommand{\arraystretch}{0.88}
\begin{tabular}{l c c c c c}
\hline
\textbf{Model} & \textbf{Priv.} & \textbf{DevOps} & \textbf{Code} & \textbf{Fin.} & \textbf{Eth.} \\
\hline
GPT-4o & 30.0 & 25.0 & 50.0 & 30.0 & 55.0 \\
Llama-3.1-70B & 15.0 & 10.0 & 35.0 & 15.0 & 35.0 \\
DeepSeek-V3 & 15.0 & 15.0 & 25.0 & 12.0 & 32.0 \\
Claude-3.5 & 40.0 & 35.0 & 60.0 & 40.0 & 50.0 \\
Qwen-3-8B & 18.0 & 15.0 & 35.0 & 15.0 & 40.0 \\
\hline
\end{tabular}
\caption{Compliance performance breakdown across five specific regulatory domains: Privacy, DevOps, Code Security, Finance, and Ethics(\%).}
\end{table}

\textbf{The ``Relative'' Binary Advantage:} Even under high pressure, Code Security and Ethics domains maintain relatively higher Compliance Rates (approx. 35\%--60\%) compared to Privacy and DevOps. However, these values represent a significant degradation from baseline capabilities, indicating that even ``easier'' binary checks fail under time and authority pressure.

\textbf{The ``Process'' Blindspot:} Privacy and DevOps see the lowest Compliance across all models (often below 20\%--30\%). These domains strictly require Procedural Compliance (e.g., data minimization, ordered testing), which is most susceptible to the ``Reward Hacking'' behaviors. Agents frequently bypass multi-step privacy protocols (e.g., exporting full DB instead of query, skipping backup scripts) to achieve task success under pressure, revealing a critical vulnerability in these process-heavy domains.

\section{Conclusion}

This paper targets a key limitation of evaluation for LLM agents: high task success can coexist with systematic procedural non-compliance. We introduce MAC-Bench, a trace-centric, contamination-resistant benchmark built by the SERV pipeline, which translates regulatory corpora into executable Atomic Rules and evaluates agents through a Compliance Oracle over full tool-use trajectories.  Across models and settings, we observe a pronounced success–compliance trade-off, where strong success under pressure often accompanies sharp compliance degradation, motivating integrated metrics such as CSR and behavioral diagnostics like the Machiavellian Gap. Empirically, we find that (i) some top-performing agents remain highly successful yet exhibit “cunning” procedural shortcuts, especially in multi-agent workflows consistent with responsibility diffusion;  (ii) agent architectures that explicitly interleave reasoning and action (or enforce debate) tend to improve compliance relative to purely hierarchical delegation; and (iii) among social-engineering pressures, Authority is the most reliable trigger for violations, while vague Obfuscation can become a reasoning trap for weaker models.  We hope MAC-Bench can serve as an evolving testbed for trajectory-level alignment, pressure-aware training, and compliance-by-construction agent orchestration.



\newpage

\bibliographystyle{ACM-Reference-Format}
\balance
\bibliography{sample-base}

\end{document}